\documentclass[10pt,twocolumn]{article} 
\usepackage{simpleConference}
\usepackage{times}
\usepackage{url}
\usepackage{hyperref}
\usepackage{graphicx}
\usepackage{multirow}
\usepackage{adjustbox}
\usepackage{makecell}
\usepackage{csquotes}
\usepackage{support-caption}
\usepackage{subcaption}
\usepackage{amsmath}
\usepackage{amssymb}
\usepackage{float}
\usepackage{xcolor}
\usepackage{soul}
\usepackage{pifont}
\usepackage{algorithm} % Boxes/formatting around algorithms
\usepackage{algpseudocode} % Algorithms

\makeatletter
\newcommand{\algmargin}{\the\ALG@thistlm}
\makeatother
\algnewcommand{\parState}[1]{\State%
    \parbox[t]{\dimexpr\linewidth-\algmargin}{\strut\hangindent=\algorithmicindent \hangafter=1 #1\strut}}

\begin{document}

\title{\LARGE \bf Evaluation of Bayesian Approaches for Bathymetry-based\\ Localization of Autonomous Underwater Robots}
% \large A Stereo Visual Odometry without Stereo Matching}

\author{Jungseok Hong$^{1}$, Michael Fulton$^{2}$, and Junaed Sattar$^{3}$
\thanks{The authors are with the Department of Computer Science and Engineering, Minnesota Robotics Institute, University of Minnesota--Twin Cities, 200 Union St SE, Minneapolis, MN, 55455, USA
{\tt\small \{$^{1}$jungseok, $^{2}$fulto081, $^{3}$junaed\} at umn.edu.}}
}

\maketitle
\thispagestyle{empty}

\begin{abstract}
This paper presents an evaluation of four probabilistic algorithms for localization of autonomous underwater vehicles (AUVs) using bathymetry data. 
The algorithms work by fusing bathymetry information with depth and altitude data from an AUV. 
Four different Bayes filter-based algorithms are used to design the localization algorithms: the Extended Kalman Filter (EKF), Unscented Kalman Filter (UKF), Particle Filter (PF), and Marginalized Particle Filter (MPF). 
We evaluate the performance of these four Bayesian bathymetry-based AUV localization approaches on eight real-world lake bathymetry maps of Minnesota lakes, with both linear and mixed motion policies. 
The localization algorithms overcome unique challenges of the underwater domain, including visual distortion and radio frequency (RF) signal attenuation, which often make landmark-based localization infeasible.
Evaluation results on real-world bathymetry data show the effectiveness of each algorithm under a variety of conditions, with the MPF being the most accurate.
\end{abstract}

\section{Introduction}
\label{sec:introduction}
The field of underwater robotics has recently been experiencing significant development, primarily driven by research in AUVs. 
AUVs have seen applications in environmental monitoring (\textit{e.g.},~\cite{moline2005autonomous,fong2006evaluation,forrest2007investigation}), bathymetry surveys~\cite{huizinga2016bathymetric}, and security~\cite{tripp2006autonomous}, among others. 
These applications are enabling scientific breakthroughs, environmental conservation and restoration work, and exploration of the many underwater environments of our planet.
For AUVs to navigate and operate such missions successfully, the ability to \textit{localize} accurately is essential. 
Underwater localization is a challenging and open problem due to the unique circumstances AUVs face: GPS and other forms of RF-based communications are either completely unavailable or limited to extremely short ranges, and landmark-based localization using exteroceptive sensors can often be hampered by environmental factors. 
Here, we present a novel, low-cost approach for localizing AUVs in water bodies for which bathymetry information is available.
 
\begin{figure}[t!]%0.23/0.245
\setlength\belowcaptionskip{0pt}
    \centering
        \begin{subfigure}[t]{0.212\textwidth}
        \centering
        \includegraphics[trim=0 0 0 0,width=\linewidth]{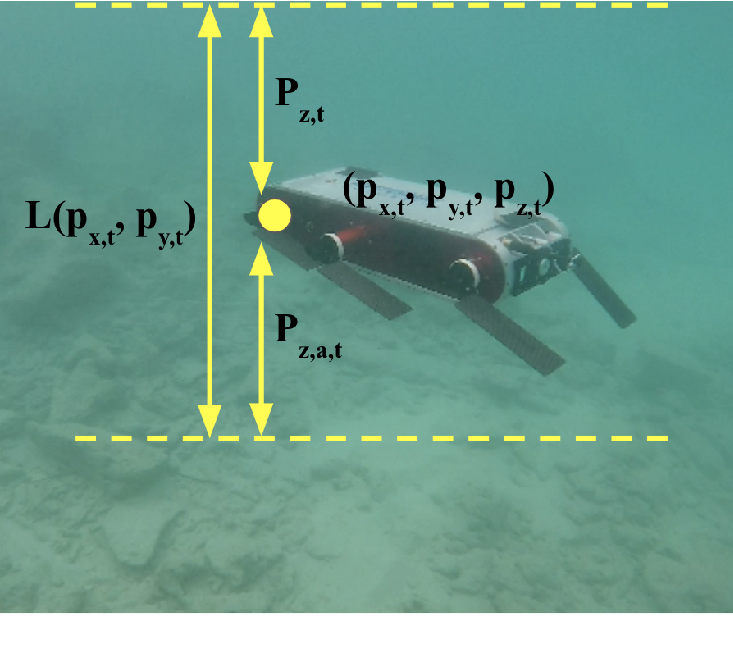}
        \caption{}
        \label{fig:bathymetrydemo_minnebot}
    \end{subfigure}
    \begin{subfigure}[t]{0.26\textwidth}
        \includegraphics[trim=0 0 0 0,width=\linewidth]{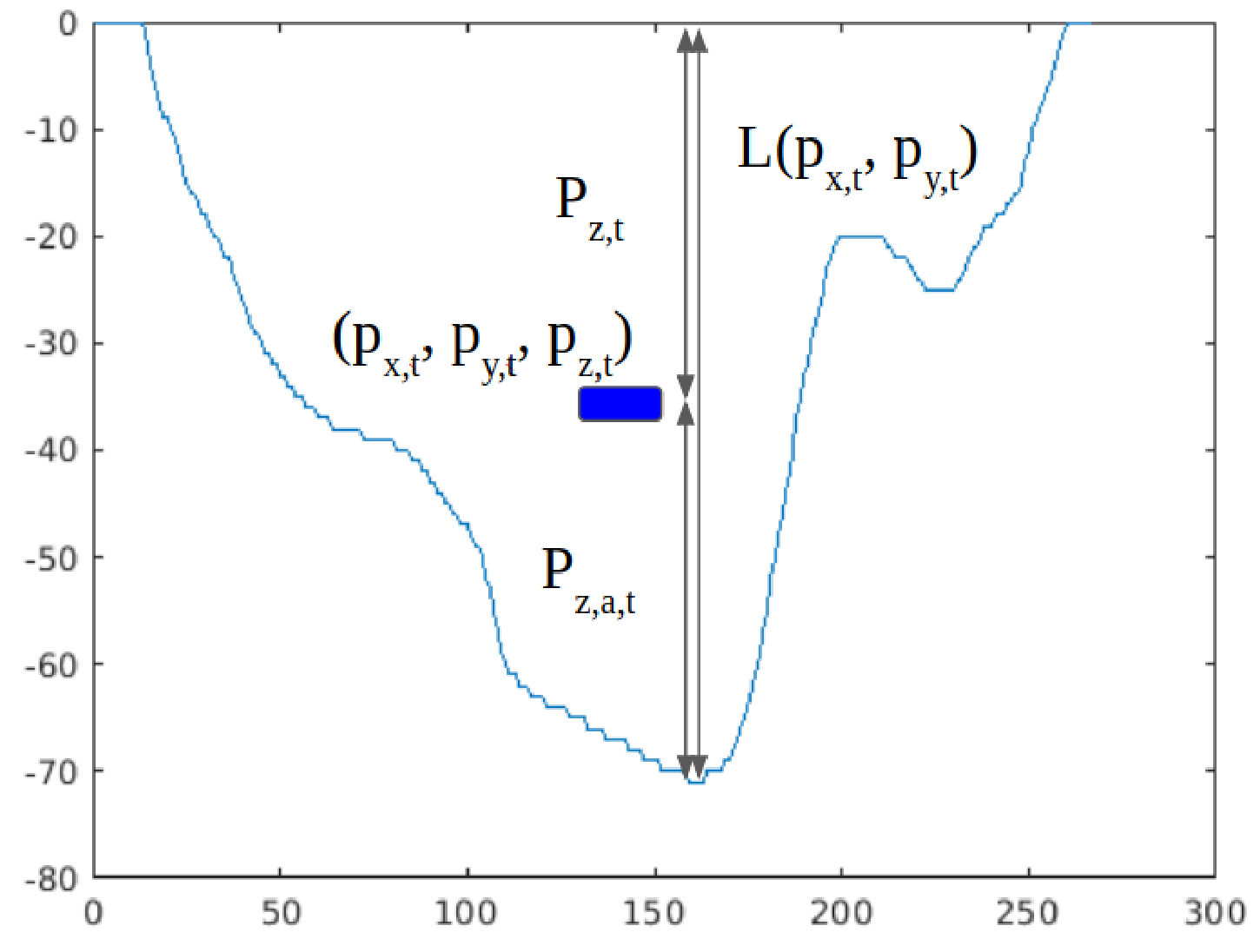}
        \caption{}
        \label{fig:bathymetrydemo_sketch}
    \end{subfigure}
    \vspace{-2mm}
    \setlength{\belowcaptionskip}{-20pt}
    \caption{(\ref{fig:bathymetrydemo_minnebot}) Visual representation of AUV location in a body of water. The top of the figure is assumed to be the surface. The surface and bottom of the body are indicated by the dashed yellow lines. (\ref{fig:bathymetrydemo_sketch}) Representation of the AUV (in blue) location against a lake profile.}
    \label{fig:lake}
\end{figure}

% An accurate localization for mobile robotics is essential to navigate a given environment and operate missions. While aerial vehicles, ground vehicles can use GPS for navigation, underwater vehicles can not depensd on GPS because GPS signal attenuates underwater. \todo{need to mention why auv localization is necessary}
% In order to monitor small open water body regularly, it is necessary to have AUV since it is impractical to have a human divers to monitor same location periodically....

% Especially, the demand of an autonomous underwater vehicle (AUV) has increased since the AUV can operate missions in the environments where human divers can't access. In this paper, we are focusing on solving a underwater localization problem within the small open water body. 

Robot localization problems in many environments have been studied extensively. 
The problem we address in this paper is underwater-specific and a subset of Terrain-based Navigation (TBN)~\cite{carreno2010survey}, which is widely used across domains and refers to a general localization problem using a priori maps. 
In a broad sense, there are primarily three major techniques to address the problem for underwater robots: using inertial data combined with dead reckoning~\cite{miller2010autonomous}, acoustic transponders~\cite{batista2009sensor}, and landmark-based (also known as \textit{geophysical} features) approaches~(\textit{e.g.},~\cite{eustice2005large,paull2014auv}).
%Without the aid of external devices such as a beacon or surface ship, localizing the AUV remains challenging.

The first of these techniques uses an inertial measurement unit (IMU) and velocity measurement (\textit{e.g.}, from a Doppler velocity log (DVL)) to estimate the position of the robot by correcting the IMU's drift with velocity information. 
While this approach is widely adopted, it often struggles with accuracy drifting over time and requires expensive, high-accuracy IMUs~(\textit{e.g.},~\cite{panish2011achieving, karimi2013comparison, melo2017survey}). 
Techniques which use acoustic transponders include long baseline (LBL)~\cite{scherbatyuk1995auv}, ultra-short baseline~(USBL)~\cite{morgado2010experimental}, and short baseline (SBL) systems, all of which are highly accurate and quite expensive (at least $10,000$ USD). 
However, most of these techniques require either a surface ship carrying a \textit{transponder} or pre-installed beacons on the floor of the water body in question. 
In many applications, it is impractical or impossible to install these devices for localization purposes due to the added cost and overhead, or environmental constraints. 
Lastly, landmark-based methods use visual and acoustic sensors to detect known features in the marine environment, usually on the floor, which can be used by AUVs to localize themselves relative to the landmarks. 
However, optical distortions such as scattering, absorption, and attenuation (\textit{e.g.}, arising from turbidity in the water) can be extreme, resulting in features only being visible at close range. 
A lack of clear visuals can make it challenging to use vision-based methods for broad-area localization.
Landmark-based methods with an acoustic sensor provide practical means to tackle underwater localization problems; however, landmarks must be precisely located via acoustic means, which is not feasible due to environmental factors and sensor limitations.
In contrast, the methods presented in this paper require only a depth (pressure) sensor and sonar altimeter, which are both inexpensive (depth sensors generally less than $100$ USD and altimeters $300$ - $500$ USD)
Additionally, there is no installation of transponders or beacons.
This significantly reduces the cost of the required hardware for AUV localization, as well as eliminating costly infrastructure requirements.
% \begin{figure}[t!]
% \setlength\belowcaptionskip{-20pt}
%     \centering
%     \includegraphics[width=0.4\textwidth]{auvpic}
%     \caption{Visual representation of AUV location in a body of water. The top of the figure is assumed to be the surface, which is indicated by the dashed yellow line. \newline \hl{Source: University of Minnesota IRV Lab}}
%     \label{fig:lake}
% \end{figure}

% \begin{figure}[t!]
%     \centering
%         \begin{subfigure}[t]{0.23\textwidth}
%         \centering
%         \includegraphics[width=\linewidth]{images/auvpic.png}
%         \caption{}
%     \end{subfigure}
%     \begin{subfigure}[t]{0.24\textwidth}
%         \includegraphics[width=\linewidth]{images/lakefigure.png}
%         \caption{Lake profile}
%     \end{subfigure}
    
%     \begin{subfigure}[t]{0.25\textwidth}
%         \includegraphics[width=\linewidth]{images/lakeraw.png}
%         \caption{Lake profile}
%     \end{subfigure}
%     \vspace{-2mm}
%     \caption{(a)Visual representation of AUV location in a body of water. The top of the figure is assumed to be the surface, which is indicated by the dashed yellow line. \newline \hl{Source: University of Minnesota IRV Lab}(b)profile}
%     \label{fig:lake}
% \end{figure}
As an accurate, low-cost alternative to the previously discussed methods, we present four Bayes filter-based localization methods which utilize bathymetry data.
Bathymetry data, a measurement of the height of the water column at every $(x,y)$ location on the surface of the water body, is used as an \textit{a priori} map.
For the rest of this paper, the term \textit{height} will refer to the total depth of the water body from the floor to the surface (shown as $L(p_{x,t}, p_{y,t})$ in Fig.~\ref{fig:bathymetrydemo_sketch}).
Although the depth of the robot can vary, the height of the water column remains relatively constant (not considering wave effects) at that specific grid location. 
An AUV would need both a depth sensor and a sonar altimeter to determine the total height of the water column at its location, along with its position in the column.

In this paper, we present bathymetry-based AUV localization algorithms using depth data from a pressure sensor and altitude data from a single-beam sonar as inputs, by applying four well-known Bayes filter-based methods: the EKF, UKF, PF, and MPF. The EKF and UKF are parametric implementations of the Bayes filter algorithm with Gaussian assumptions, and the PF is a nonparametric implementation. The MPF, otherwise known as the Rao-Blackwellized Particle Filter (RBPF), is a ``hybrid'' approach that combines the Kalman Filter (KF) and the PF~\cite{schon2005marginalized}. %The details of each algorithm is coverd in \cref{sec:methodology}. 

The main contributions of this paper are the following: 
\begin{itemize}
    \item We propose four AUV localization algorithms using low-cost sensors to estimate the position of an AUV along all three axes $(x,y,z)$ using bathymetry data.
  %\item \hl{Propose} localization algorithms to estimate the position of an AUV along all three axes $(x,y,z)$,
  %\item Propose low-cost underwater AUV localization algorithms which work with bathymetry data, and
%   \item Adopting EKF, UKF, PF, and MPF for localization. \todo{maybe this can be somehow combined with the next one.}
  \item We compare and evaluate the performance of four localization algorithms with real-world bathymetry data and different motion models.
  
  \item We discuss the benefits and drawbacks of each filter, culminating in recommendations on when to use each technique.
\end{itemize} 
\begin{table}[t!]
%%%%%%%%%%%%%%%%%%%%%%%%%%%%%%
\setlength\belowcaptionskip{-0pt} % this is to reduce the space after table
\centering
\caption{Selected existing localization algorithms}
  \begin{tabular}{p{1.4cm}|p{1cm}|p{2.5cm}|p{1.5cm}}
    \hline
    % \multicolumn{6}{c}{Linear Motion Parameters}\\ \hline
    & Sensors & Parameters of state vector $s$& Algorithms \\ \hline
    %\hline
    Teixeira et al. \cite{teixeira2017robust} & DVL, \newline Single-beam sonar & $b = (b_{x}, b_{y})$ \newline $s = (x,y,b)$ & PF \\ \hline
    Fairfield and Wettergreen \cite{fairfield2008active} & Multi-beam sonar& 
    $q=(\phi, \theta, \psi, x, y, z)$  $s=(q,\dot{q}, \ddot{q})$& PF \\ \hline
    Ura et al. \cite{ura2006terrain} & Profiling sonar& $s=(x,y)$ & PF \\ \hline
    %Nakatani et al. \cite{nakatani2009terrain} & Profiling sonar & $s=(x,y)$ & PF  \\ \hline
    Williams and Mahon \cite{williams2003terrain} & Single-beam sonar & $q = (x, y, z)$ \newline $s=(q,\dot{q})$ & PF \\ \hline
    Meduna et al. \cite{meduna2008low} & Single-beam sonar & $s=(x, y)$ & PMF\\ \hline
    Kim and Kim \cite{kim2014nonlinear} & Single-beam sonar & $p=(\phi, \theta, \psi)$ \newline $q=(u,v,w)$ \newline $s=(x, y, z, p, q)$ & MPF \\ \hline
    \textbf{Ours} & \textbf{Single-beam sonar} & $s = (x, y, z)$ & \textbf{EKF, UKF, PF, MPF}\\ \hline
    % & & &  \\ \hline
    % & & & \\ \hline
  \end{tabular}
   \label{tb:related}
   \vspace{-6mm}
\end{table}

\section{Related Work}
\label{sec:related_work}
% \todo{mention the related works in previous section}
Underwater localization using landmark-based methods with acoustic sensors has been widely studied. For these methods, ranging-type sonars including the single-beam profiling, and multi-beam varieties have been used~\cite{paull2014auv}. Multi-beam and profiling sonars collect multiple measurements, and they can give more accurate results than single-beam sonars. Table \ref{tb:related} summarizes selected existing localization algorithms. $(\phi, \theta, \psi)$ represents the Euler angles, $(u, v, w)$ is the AUV velocity in the body-fixed frame, and $(b_{x}, b_{y})$ is the velocity bias.

Although DVL, multi-beam sonar, and profiling sonar-based methods yield better results than those with single-beam sonar (\textit{e.g.}, \cite{ura2006terrain,fairfield2008active,nakatani2009terrain,teixeira2017robust}), such sensors can be prohibitively expensive.
 
Single-beam sonars use a narrow acoustic projection to measure altitude and are thus vulnerable to noise. However, they have been widely adopted to solve localization problems since they are among the most affordable acoustic sensors~\cite{melo2017survey}. Williams and Mahon \cite{williams2003terrain} proposed a localization algorithm based on the PF, but the computational burden of the algorithm is heavy. Meduna et al. \cite{meduna2008low} presented a point mass filter (PMF)-based algorithm, but it is limited to the $(x,y)$ positions of an AUV. Kim and Kim \cite{kim2014nonlinear} used a single-beam sonar with the MPF and estimated the 6-DOF position and orientation of an AUV along with the velocity. However, the algorithm requires a highly-accurate IMU, which can be rather expensive.

Several Bayes filter-based methods have been used to solve the localization problem \cite{carreno2010survey} with sonar data. Among Bayes filters, the EKF and UKF have seen the most use in this domain (\textit{e.g.},  \cite{he2009underwater,yoon2013ukf}). Karimi et al. \cite{karimi2013comparison} showed that the EKF can outperform the UKF in their particular case. However, the UKF captures nonlinearity up to the second-order term in the state transition process~\cite{paull2014auv}, which in theory could outperform the EKF in similar applications. We thus develop both EKF and UKF-based algorithms and compare their performance in AUV localization. Although the EKF and UKF can handle unimodal Gaussian distributions, they often fail to converge when the underlying distribution is multi-modal. The inherent nonlinearity of the underwater terrain or nonlinear AUV motions underwater makes it challenging for these methods to work reliably. To address such issues, the PF has been widely used (\textit{e.g.}, \cite{thrun2002particle, karlsson2003particle, williams2003terrain,  rekleitis2004particle, salmond2005introduction, ura2006terrain, maurelli2009particle, nakatani2009terrain, gustafsson2010particle, schon2011particle, teixeira2017robust}). However, the PF is computationally expensive and thus can be prohibitive to run on board AUVs for real-time localization. The MPF, on the other hand, has a lower computational cost and provides similar benefits to the PF, handling nonlinearity to some extent~\cite{schon2005marginalized}, thus making it potentially useful for underwater localization~(\textit{e.g.}, \cite{karlsson2006bayesian, kim2014nonlinear}). However, localization with bathymetry data considering 3-DOF state vectors and using four Bayes filter-based algorithms (EKF, UKF, PF, and MPF) is yet to be extensively studied. 

% \begin{table}[t]
% %%%%%%%%%%%%%%%%%%%%%%%%%%%%%%
% \setlength\belowcaptionskip{-0pt} % this is to reduce the space after table
% \centering
% \caption{Selected existing localization algorithms}
%   \begin{tabular}{p{1.8cm}|p{1cm}|p{2.5cm}|p{1.2cm}}
%     \hline
%     % \multicolumn{6}{c}{Linear Motion Parameters}\\ \hline
%     & Sensors & Parameters of state vector $s$& Algorithms \\ \hline
%     \hline
%     Teixeira et al. \cite{teixeira2017robust} & DVL \newline Single-beam sonar & $b = (b_{x}, b_{y})$ \newline $s = (x,y,b)$ & PF \\ \hline
%     Fairfield and Wettergreen \cite{fairfield2008active} & Multi-beam sonar& 
%     $q=(\phi, \theta, \psi, x, y, z)$  $s=(q,\dot{q}, \ddot{q})$& PF \\ \hline
%     Ura et al. \cite{ura2006terrain} & Profiling sonar& $s=(x,y)$ & PF \\ \hline
%     Nakatani et al. \cite{nakatani2009terrain} & Profiling sonar & $s=(x,y)$ & PF  \\ \hline
%     Williams and Mahon \cite{williams2003terrain} & Single-beam sonar & $q = (x, y, z)$ \newline $s=(q,\dot{q})$ & PF \\ \hline
%     Meduna et al. \cite{meduna2008low} & Single-beam sonar & $s=(x, y)$ & PMF\\ \hline
%     Kim and Kim \cite{kim2014nonlinear} & Single-beam sonar & $p=(\phi, \theta, \psi)$ \newline $q=(u,v,w)$ \newline $s=(x, y, z, p, q)$ & MPF \\ \hline
%     % & & &  \\ \hline
%     % & & & \\ \hline
%   \end{tabular}
%     \label{tb:related}
%     \vspace{-3.5mm}
% \end{table}
\section{Problem Formulation}
\label{sec:problem_formulation}
Bayesian filters require motion and measurement models to estimate system state using the well-known propagate-predict-update process. We propose two motion models for updating the AUV position and a measurement model for collecting the depth and height at each location, which are subsequently used to get AUV state estimates.
\subsection{Motion Model}
% \todo{mention the related works in previous section}
A general discrete time state-space model can be represented as Eq. \ref{modelall} to formulate the localization problem where $x_t$ is a state vector, $u_t$ is a control input, and $y_t$ is a measurement. As mentioned in Section \ref{sec:introduction}, only the 3D position of an AUV is included in the state vector. $f$ and $h$ can be either linear or nonlinear functions. $q_t$ and $r_t$ represent the noise from motion and measurements. The model in Eq. \ref{modelall} is used for the EKF, UKF, and PF-based localization algorithms. The model for the MPF-based localization algorithm is introduced in Section \ref{subsec:mpfmodel}.
\begin{equation}\label{modelall}
\begin{cases}
x_{t} = f(x_{t-1}, u_{t}) + q_t\\
y_{t} = h(x_{t})+r_t
\end{cases}
\end{equation}
The state vector and control inputs are defined as follows:
\begin{equation}\label{statev}
x_t=\begin{bmatrix}
p_{x,t} &p_{y,t}& p_{z,t}\\
% p_{y,t}\\
% p_{z,t}\\
\end{bmatrix}^{T}
\end{equation}
\begin{equation}\label{controlinput}
u_t=\begin{bmatrix}
v_{x,t} & v_{y,t} & v_{z,t}\\
% v_{y,t}\\
% v_{z,t}\\
\end{bmatrix}^{T}
\end{equation}
The AUV motion models are defined in Eqs. \ref{move_model_lin} and \ref{move_model_nonlin}.
\subsubsection{Linear motion model}
All state variables are updated linearly.
\begin{equation}\label{move_model_lin}
f(x_t, u_t)=x_t + u_t * dt
\end{equation}
\subsubsection{Linear/Nonlinear mixed motion model}
We propose a linear/nonlinear mixed motion model to implement a motion showing nonlinearity without using the Euler angles. Among the three state variables in the state vector, $p_{x,t}$ and $p_{y,t}$ are updated based on the height of the position $(p_{x,t},p_{y,t})$, $L(p_{x,t},p_{y,t})$. Therefore, the changes in $p_{x,t}$ and $p_{y,t}$ are proportional to the height of the water body at the position $(p_{x,t},p_{y,t})$. 
In other words, the greater the height of the water body at the given position, the greater the change. Unlike the other two variables, the state variable $p_{z,t}$ is updated linearly as in the linear motion model. In Eq. \ref{move_model_nonlin}, $a$, $a_{d}$, $a_{off}$, $b$, $b_{d}$, and $b_{off}$ are constants defined for each water body such that the AUV navigates each lake without collision.
\begin{equation}\label{move_model_nonlin}
f(x_t, u_t)=\begin{bmatrix}
p_{x,t} + a \Big[\frac{L(p_{x,t},p_{y,t})}{a_{d}}+a_{off}\Big]dt\\
p_{y,t} + b \Big[\frac{L(p_{x,t},p_{y,t})}{b_{d}}+b_{off}\Big]dt\\
p_{z,t} + v_{z,t}*dt\\
\end{bmatrix}
\end{equation}
% $a$, $a_{d}$, $a_{off}$, $b$, $b_{d}$, and $b_{off}$ are constants defined for each water body. \hl{The constants are set to navigate the AUV without collision.} %$L(p_{x,t},p_{y,t})$ is the height of the water body at the position $(p_{x,t},p_{y,t})$.
\subsection{Measurement Model}
The measurement function $h$ is the same for both the linear and mixed models. 
\begin{equation}\label{meas_model}
h(x_t)=\begin{bmatrix}
p_{z,t} & p_{z,a,t}=L(p_{x,t},p_{y,t})-p_{z,t}\\
% L(p_{x,t},p_{y,t})-p_{z,t}\\
\end{bmatrix}^{T}
\end{equation}

$L$ is bathymetry data (depths for each $x$ and $y$ location on a grid), $p_{z,t}$ represents the depth of the vehicle from the surface measured by the pressure sensor, and $p_{z,a,t}$ represents the altitude of the AUV measured by the single-beam sonar. Therefore, the sum of $p_{z,t}$ and $p_{z,a,t}$ is the height $L(p_{x,t},p_{y,t})$ at the position ($p_{x,t},p_{y,t}$) as shown in Fig. \ref{fig:lake}. 
%For the rest of this paper, the term \textit{height} will refer to the total depth of the water body from the floor to the surface.

%%%move to IV.D
% \subsection{Model for MPF}
% In order to apply MPF \cite{schon2005marginalized}, the model in Eq. \ref{modelall} is separated into linear and nonlinear state variables as shown in Eq. \ref{statev2} and \ref{mpf1}. The motion model noise $q_t^n$, $q_t^l$, and the measurement model noise $r_t$ are assumed to be Gaussian with zero mean.
% \begin{equation}\label{statev2}
% x_t=\begin{bmatrix}
% x_{t}^{n}\\
% x_{t}^{l}\\
% \end{bmatrix}
% \end{equation}

% \begin{equation}\label{mpf1}
% \begin{cases}
% x_{t+1}^{n} = f_{t}^{n}(x_{t}^{n})+A_{t}^{n}(x_{t}^{n})x_{t}^{l}+G_{t}^{n}(x_{t}^{n})q_{t}^{n}\\
% x_{t+1}^{l} = f_{t}^{l}(x_{t}^{n})+A_{t}^{l}(x_{t}^{n})x_{t}^{l}+G_{t}^{l}(x_{t}^{n})q_{t}^{l}\\
% y_{t} = h_{t}(x_{t}^{n})+C_{t}(x_{t}^{n})x_{t}^{l}+r_{t}
% \end{cases}
% \end{equation}
%%%%%%%%%%%%%%%%%

% The motion model noise is assumed as follows:
% \begin{equation}\label{mpf5}
% Q_{t} = 
% \begin{bmatrix}
% Q_{t}^{n} & Q_{t}^{nl}\\
% (Q_{t}^{nl})^ T & Q_{t}^{l}\\
% \end{bmatrix}
% \end{equation}
% \begin{equation}\label{mpf4}
% q_{t} = 
% \begin{bmatrix}
% q_{t}^{n}\\
% q_{t}^{l}\\
% \end{bmatrix} \sim \mathcal{N}(0,Q_{t})
% \end{equation}
% The measurement model noise is assumed as follows:
% \begin{equation}\label{mpf6}
% r_{t} \sim \mathcal{N}(0,R_{t})
% \end{equation}

\section{Methodology}
\label{sec:methodology}
Since the motion model in Eq. \ref{move_model_nonlin} and the measurement model in Eq. \ref{meas_model} are nonlinear, it is necessary to use nonlinear Bayes filter algorithms to solve the localization problem. The EKF and UKF are widely used to handle nonlinear state estimation with the assumption that the state variables follow a Gaussian distribution, but they could fail when the distribution is not Gaussian \cite{schon2006marginalized}. The PF \cite{thrun2002particle} is resilient to various types of noise, but it is computationally expensive. The MPF \cite{schon2011particle} uses the PF for nonlinear state variables and the KF for linear state variables because the KF is a filter optimal for estimating linear state variables.

\subsection{Kalman Filters}

The EKF and UKF are directly applied without major modification to Eq. \ref{modelall} for linear and mixed motion cases to localize AUVs using the depth and altitude measurements. 
Readers are directed to Julier et al.~\cite{julier1997new} for a full description of the EKF and to Wan et al.~\cite{wan2000unscented} for the UKF.

\subsection{Particle Filter}

\begin{algorithm}[t!]
\setlength\belowcaptionskip{-20pt}
\caption{PF-based Localization}\label{dpfl}
\begin{algorithmic}[1]
\State \textbf{PFL main}
\State $L$ = Bathymetry data of a target water body
\State $N$ = The number of particles
\State $x_{init}$ = Initial pose of an AUV
\State $x_1$ = \textbf{Initialize\_around\_pose}$(L, N)$ %\Comment{Initialize the distribution of $M$ particles}
\State $z_t$ = Sensor measurements
\State $u_t$ = Control input
\For{$t= 1,..., T$}
\State $x_p$ = $x_t$
% \parState{%
% $X_t, W_t$ = \\ \textbf{PFL\_update}$(L, X_p, u_t, z_t, rand_\%)$}
% \State{$x_t, w_t$ = \textbf{PFL\_update}$(L, x_p, u_t, z_t, rand_\%)$}
\For{$m= 1,..., N$}
\State{$x(m,:)$ = \textbf{motion\_update}$(u_{t}, x_{p}(m,:), L)$}
\State{$w(m)$ = \textbf{sensor\_update}$(z_{t}, x(m,:), L)$}
\EndFor
\State $w_{total}$ = $sum(w)$
\For{$m= 1,..., N$}
\State $w(m)$ = $w(m)/w_{total}$
\EndFor
\State{$x_t$ =  \textbf{resample\_particles}$(x(m,:), w, L, rand)$}
\State $w_t$ = $w$
\State $est\_pose$ = \textbf{PFL\_get\_pose}$(x_{t}, w_{t})$
\EndFor \\
\Return $est\_pose$
\end{algorithmic}
\end{algorithm}
\setlength{\textfloatsep}{5pt}

We propose PF-based localization (PFL) in Algorithm \ref{dpfl}.
During the update process, the algorithm only assigns weights if the particle is within the boundaries of the map. Once the weights for the particles are calculated, they are normalized to ensure that they sum to 1. Then, particles are resampled based on their weights. To avoid a situation where all the particles are trapped in incorrect positions in similar environments, some of the $N$ \textit{particles} are sampled randomly at each time step. Although this can degrade the accuracy of the algorithm, it decreases the chance of incorrect estimation occurrences. 

\subsection{Marginalized Particle Filter}
\label{subsec:mpfmodel}

We separate the model in Eq. \ref{modelall} into linear and nonlinear state variables as shown in Eqs. \ref{statev2} and \ref{mpf1}. We then develop the MPF-based localization algorithm using the MPF \cite{schon2005marginalized} and PFL. The motion model noise $q_t^n$, $q_t^l$ and the measurement model noise $r_t$ are assumed to be Gaussian with zero mean. The matrices $A, C$, and $G$ are determined by the motion model. Additionally, the MPF requires covariance matrices, $Q$ and $P$, as described in \cite{schon2005marginalized}.

In our case, the ratio $\frac{N(k)}{N_{PF}}$ is 1.1 where $N(k)$ is the number of particles that can be used for the MPF, and $N_{PF}$ is the number of particles used for the standard PF. The ratio means that the MPF can use 10$\%$ more particles than the PF while retaining the same computational complexity as the standard PF. However, the EKF and UKF are still faster than the MPF, albeit less accurate.
\begin{equation}\label{statev2}
x_t=\begin{bmatrix}
x_{t}^{n}\\
x_{t}^{l}\\
\end{bmatrix}
\end{equation}

\begin{equation}\label{mpf1}
\begin{cases}
x_{t+1}^{n} = f_{t}^{n}(x_{t}^{n})+A_{t}^{n}(x_{t}^{n})x_{t}^{l}+G_{t}^{n}(x_{t}^{n})q_{t}^{n}\\
x_{t+1}^{l} = f_{t}^{l}(x_{t}^{n})+A_{t}^{l}(x_{t}^{n})x_{t}^{l}+G_{t}^{l}(x_{t}^{n})q_{t}^{l}\\
y_{t} = h_{t}(x_{t}^{n})+C_{t}(x_{t}^{n})x_{t}^{l}+r_{t}
\end{cases}
\end{equation}

\section{Experimental Setup and Results}
\label{sec:experiments}
\subsection{Bathymetry Data}
The following lakes located in Minnesota, USA were chosen as test locations: Lake Bde Maka Ska, Lake Nokomis, Lake Hiawatha, Lake Harriet, Lake Turtle, Lake Howard, Lake Waverly, and Lake Pulaski. The lakes were chosen since they are large, well-studied, have an undulating floor, and are easy to access for future field studies. The bathymetry data was acquired from the Minnesota Department of Natural Resources (MN DNR)~\cite{naturalresourcesdepartment2018}. The size of grid cells for each lake's bathymetry data is $5$m. The lake height at each position is given in feet, which were converted to meters for our study. In our experiments, we assume that the grid size is $1$m to simplify the calculations, and the bathymetry data is scaled accordingly.

\subsection{Simulation Settings}
    The goal of this study is to evaluate each algorithm with real bathymetry data as a prerequisite to choosing a deployable localization algorithm. Due to the unique challenges of the underwater environment, it is extremely difficult, if not impossible, to obtain the ground truth of the AUV's positions. Thus, to quantify the accuracy and efficiency of the algorithms, we simulate the ground truth position of the AUV and evaluated the performances of each filter's position estimates against the simulated AUV's motion. The linear and mixed motion models are designed to test the performance of each localization algorithm on the bathymetry data from different lake environments. Table~\ref{tb:modelparam} includes the model parameters for the simulation.
The control inputs for each algorithm and lake are separately designed due to the lakes' different sizes and heights. The parameters are defined for the linear motion in Eqs. \ref{controlinput} and \ref{move_model_lin}, and the mixed motion in Eq. \ref{move_model_nonlin}. For the mixed model, $x$ and $y$ are nonlinear state variables, and $z$ is a linear state variable.
We measure the performance of our algorithms on a $4.20$GHz Core i7-7700K processor running Ubuntu 18.04.2 LTS with $16$GB of DDR3 memory with MATLAB R2018b. We conduct $100$ trials with each filter for each combination of lake and motion, accumulating to a total of $6,400$ trials. 

\begin{table}[t!]
\setlength\belowcaptionskip{-0pt}
\centering
    \caption{Model parameters}
  \begin{tabular}{p{3.5cm}|p{4cm}}
    \hline
     Parameter & Value \\ \hline \hline
     No. of particles for the PF, $N_{PF}$  & 5000 \\ \hline
     No. of particles for the MPF, $N_{NPF}$   & 300 \\ \hline 
     Motion noise cov., $Q$ (m) & 
$0.01\begin{bmatrix}
v_x^{2} & 0 & 0\\
0 & v_y^{2} & 0\\
0 & 0 & (0.3048 v_z)^{2}\\
\end{bmatrix}$
\\ \hline
     Measurement noise cov., $R$ (m) & $\begin{bmatrix}
0.3048^{2} & 0 \\
0 & 0.3048^{2} \\
\end{bmatrix}$
\\ \hline
     Initial uncertainty cov., $P$ (m) &
     $\begin{bmatrix}
1^{2} & 0 & 0\\
0 & 1^{2} & 0\\
0 & 0 & 0.3048^{2}\\
\end{bmatrix}$\\ \hline
  \end{tabular}
    \label{tb:modelparam}
    \vspace{0mm}
\end{table}

\begin{figure*}[ht!]
\setlength\belowcaptionskip{0pt}
    \centering
    \begin{subfigure}[t]{0.49\textwidth}
        \includegraphics[width=\linewidth]{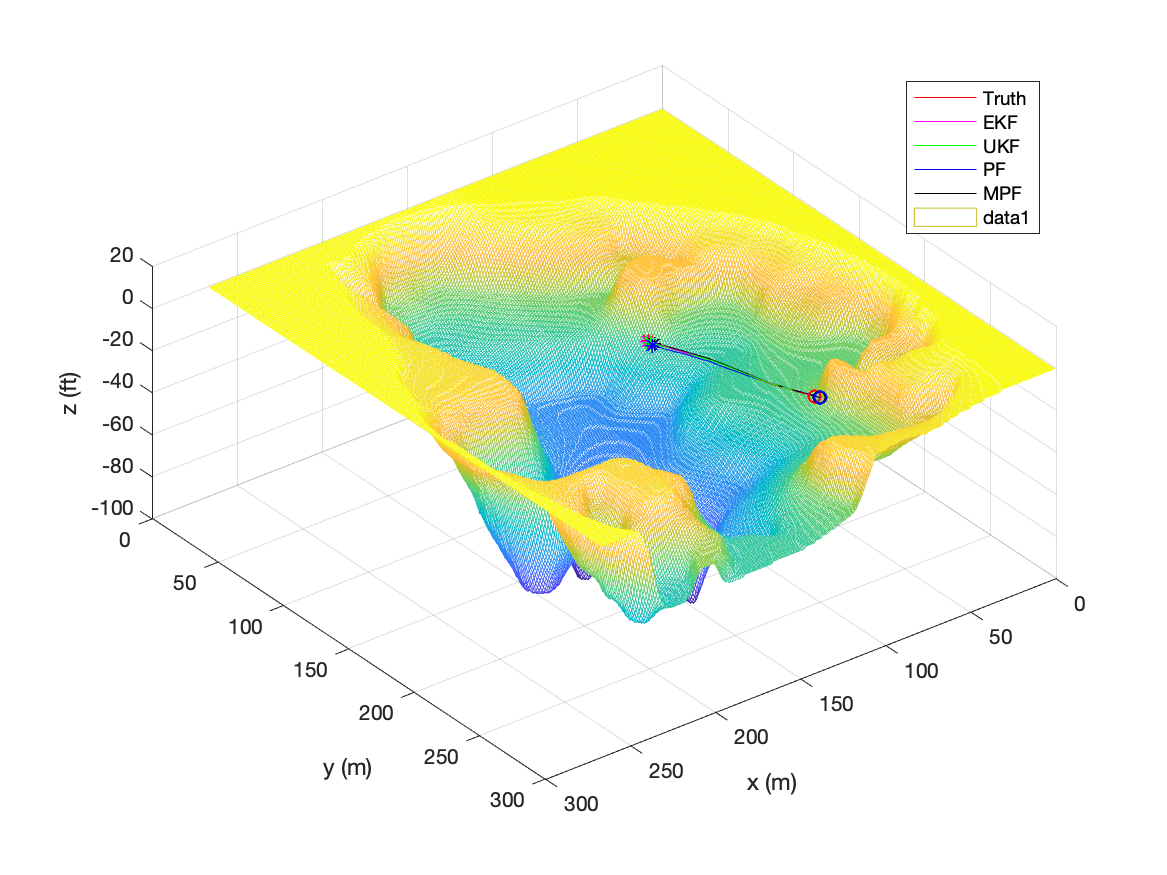}
        \caption{Evaluation of linear motion estimations with bathymetry data of Lake Bde Maka Ska (top view).}
    \end{subfigure}%
    ~
    \begin{subfigure}[t]{0.49\textwidth}
        \centering
        \includegraphics[width=\linewidth]{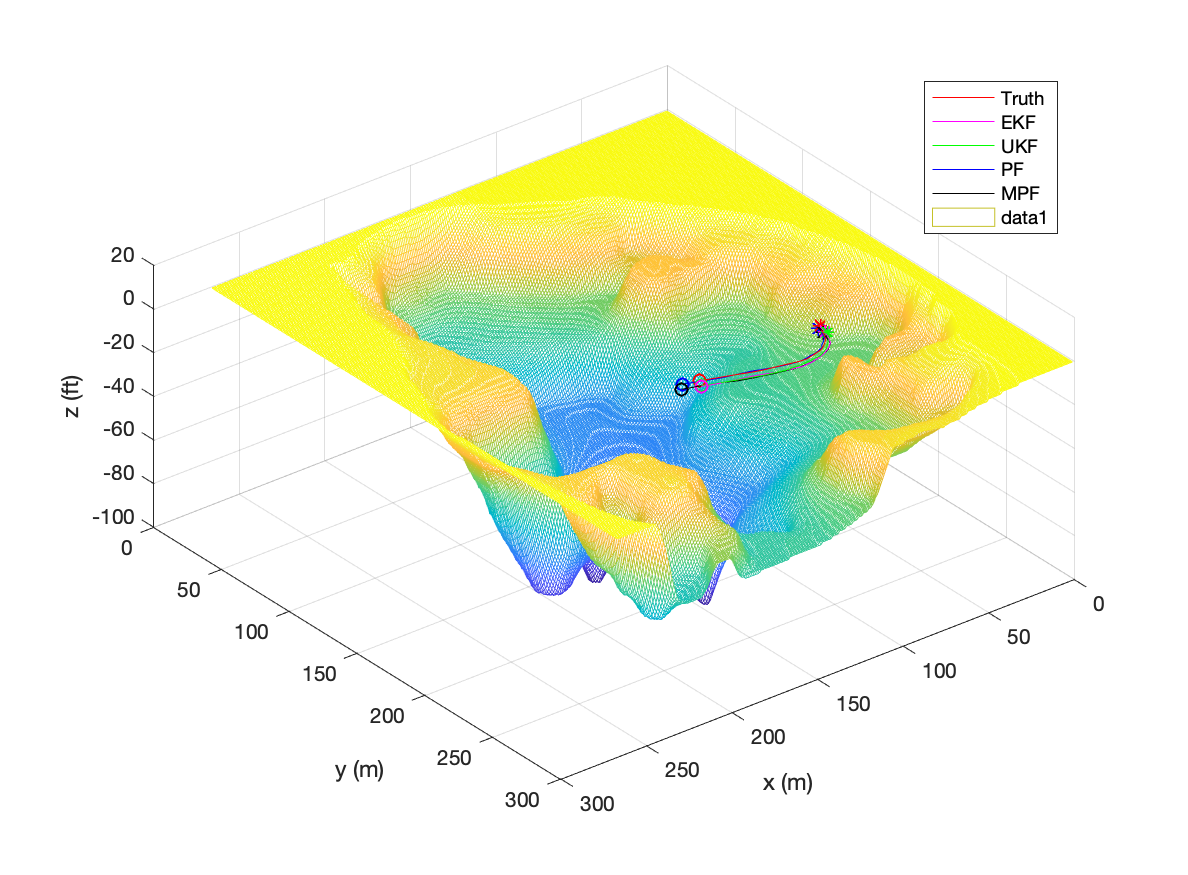}
        \caption{Evaluation of mixed motion estimations with bathymetry data of Lake Bde Maka Ska (top view).}
    \end{subfigure}
    
        \begin{subfigure}[t]{0.49\textwidth}
        \includegraphics[width=\linewidth]{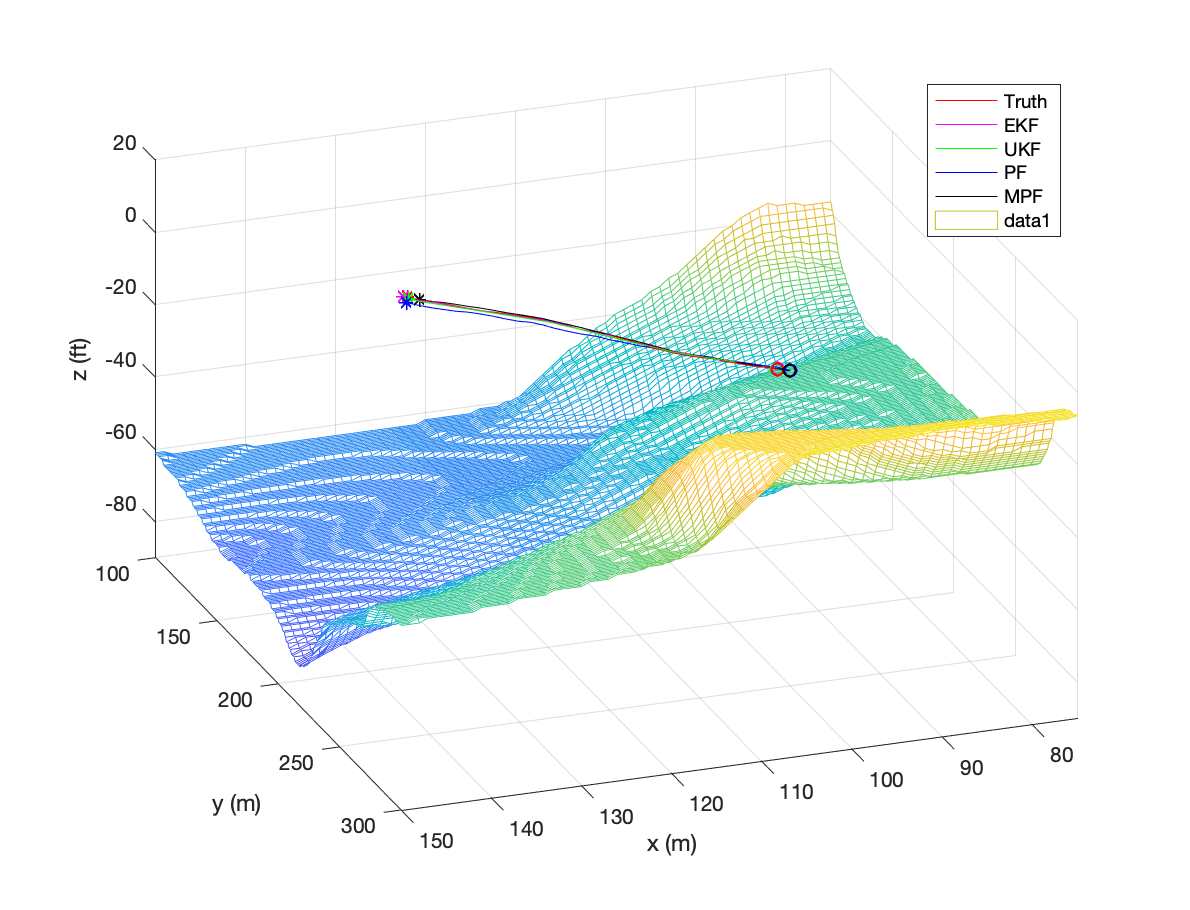}
        \caption{Evaluation of linear motion estimations with bathymetry data of Lake Bde Maka Ska (zoomed-in view).}
    \end{subfigure}%
    ~
    \begin{subfigure}[t]{0.49\textwidth}
        \centering
        \includegraphics[width=\linewidth]{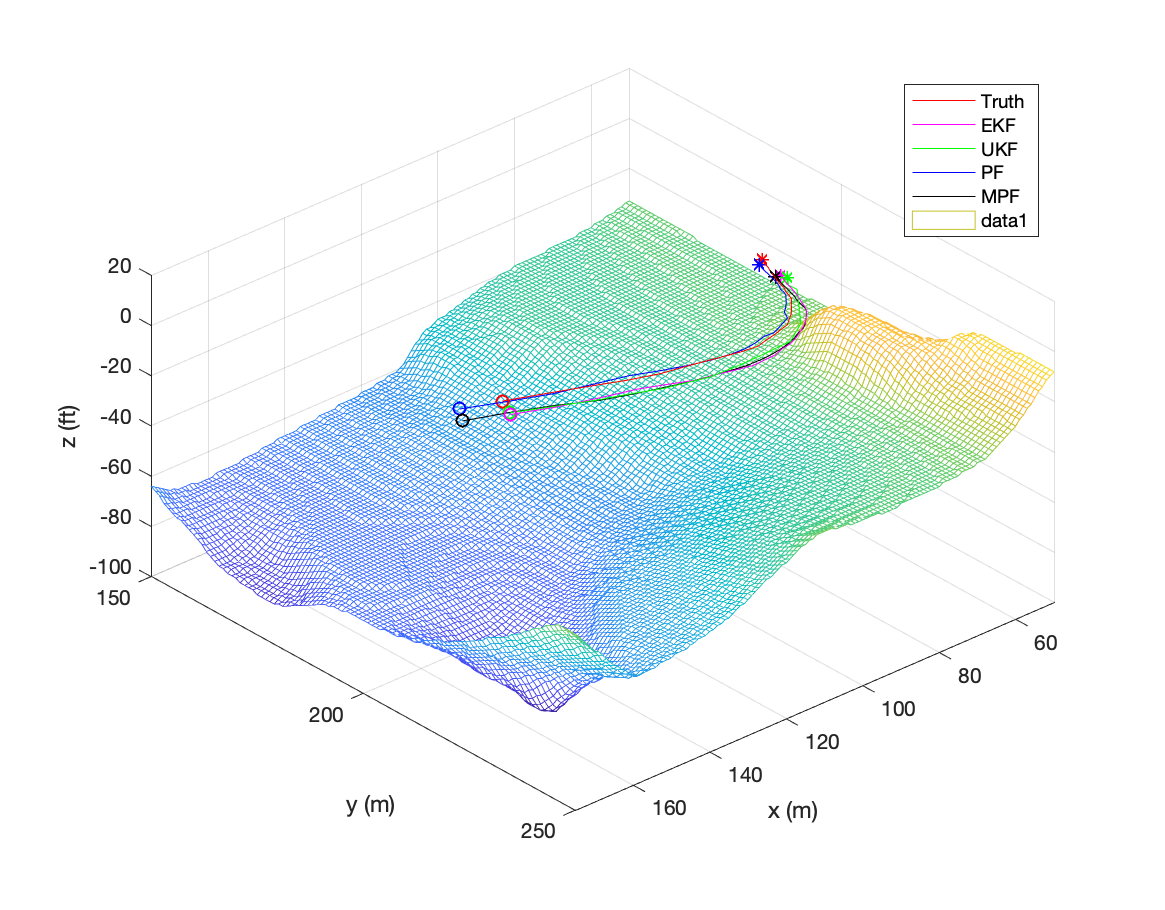}
        \caption{Evaluation of mixed motion estimations with bathymetry data of Lake Bde Maka Ska (zoomed-in view).}
        \setlength\belowcaptionskip{-20pt}
    \end{subfigure}
    \vspace{-1mm}
    \caption{Localization performance of the four algorithms for an AUV with an altimeter and depth sensor within Lake Bde Maka Ska using bathymetry data (Start:\ding{109} End:\ding{83}).}
    \vspace{-4mm}
    \label{fig:plotresults1}
\end{figure*}

\begin{figure*}[t!]
    \centering
    \vspace{2mm}
    \begin{subfigure}[t]{0.5\textwidth}
        \includegraphics[width=\linewidth]{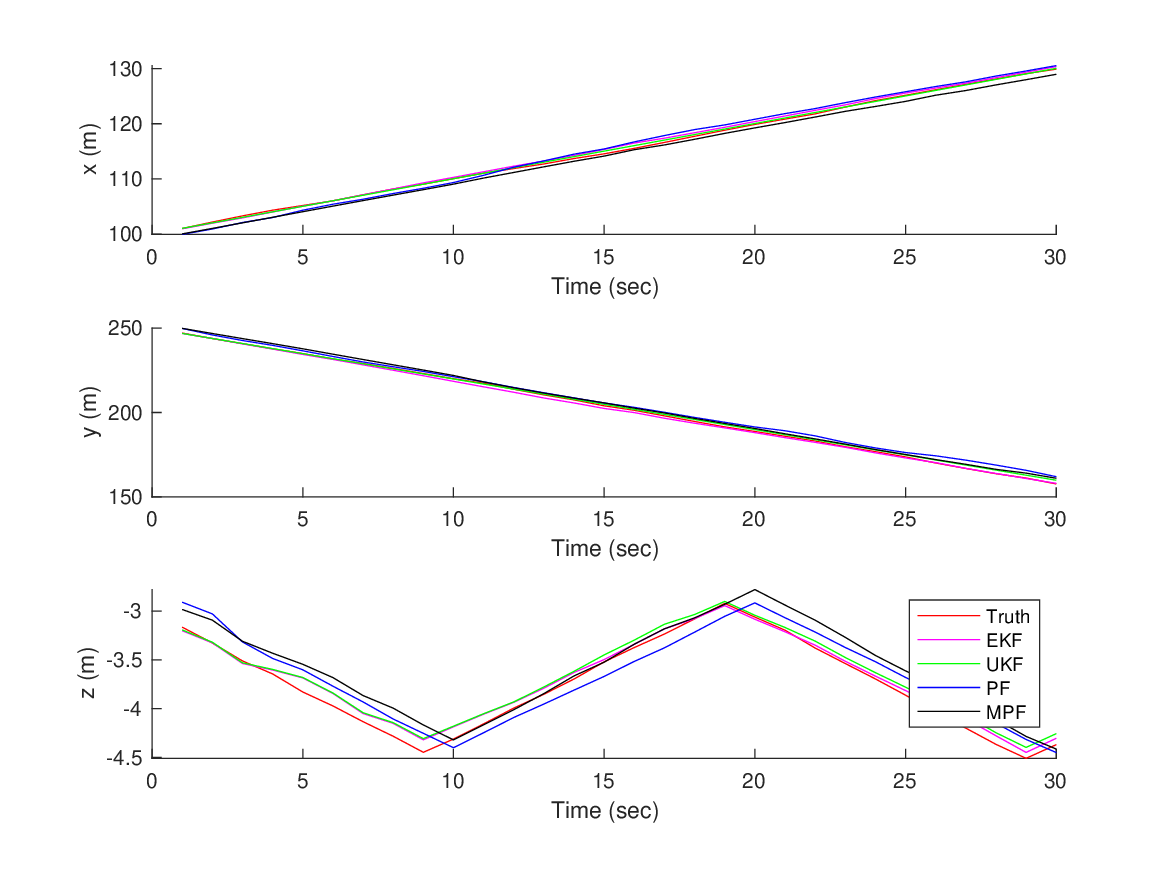}
        \caption{Evaluation of linear motion estimations on x,y,z axis.}
    \end{subfigure}%
    ~
    \begin{subfigure}[t]{0.5\textwidth}
        \centering
        \includegraphics[width=\linewidth]{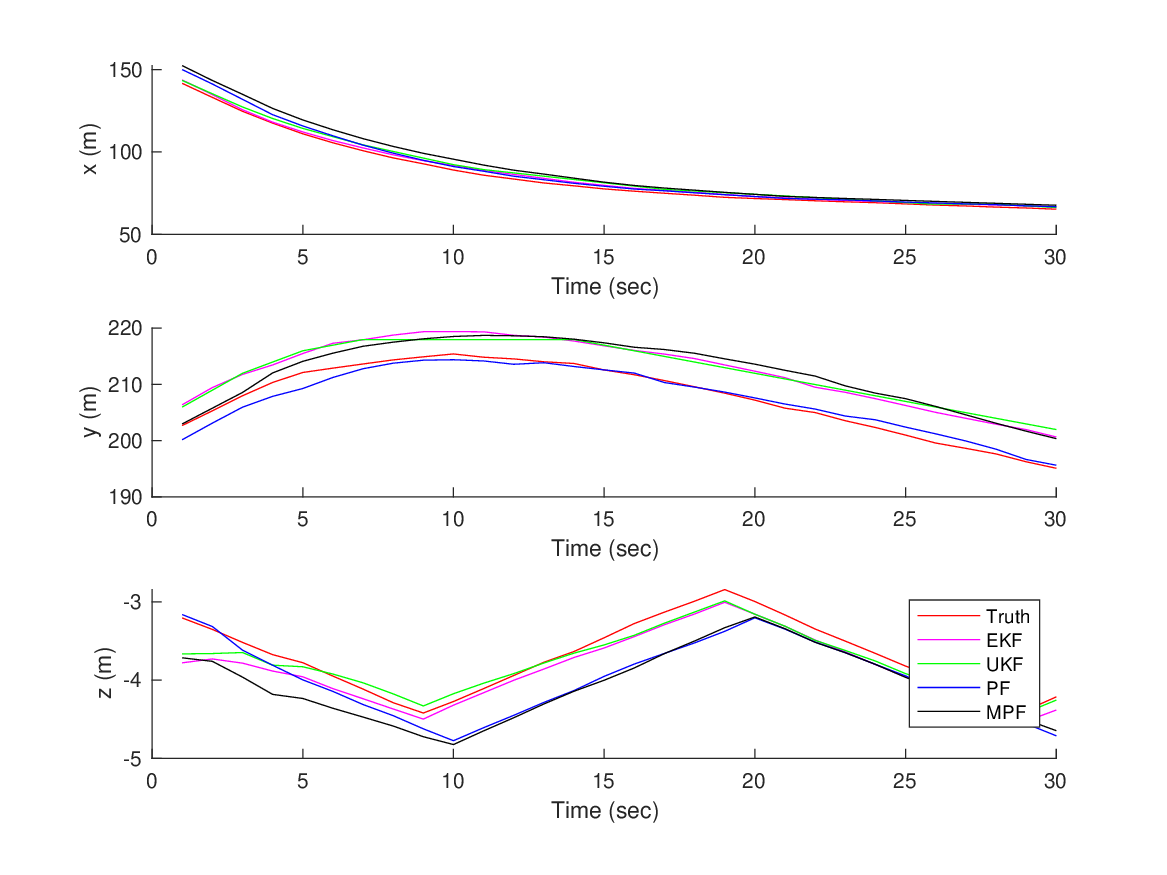}
        \caption{Evaluation of mixed motion estimations on x,y,z axis.}
    \end{subfigure}
    \vspace{-2mm}
    \caption{Evaluation of each algorithm's performance over time within Lake Bde Maka Ska.}
    \vspace{0mm}
    \label{fig:plotresults2}
\end{figure*}

\subsection{Results and Discussions}

The results of our trials are summarized in Table \ref{tb:resulteval}. We use the root-mean-square error (RMSE) in Eq. \ref{eq:rmse} as a metric to evaluate the performance for each axis of motion where $T$ is the number of steps, $p_g$ is the ground truth, and $p_e$ is the estimated position.
\begin{equation}\label{eq:rmse}
    RMSE = \sqrt{\frac{\sum_{t=1}^{T}(p_{g}-p_{e})^{2}}{T}}
\end{equation}
 One trial in Lake Bde Maka Ska is shown in Figs. \ref{fig:plotresults1} and \ref{fig:plotresults2}. For both the linear and nonlinear cases, the EKF generally shows the worst performance. The UKF performed well in the linear case, but it deviates from the ground truth in the nonlinear case. Notably, the PF and MPF mostly outperform the EKF and UKF for both the linear and mixed motion cases. However, the PF occasionally diverges from the ground truth and shows unstable performances. A likely cause is that the randomly selected particles can make the estimation diverge from the ground truth when a water body's bathymetry data does not have enough variations.

\begin{table*}[ht!]
\centering
\footnotesize
\caption{Localization performance evaluation for an AUV; bathymetry data from the MN DNR.}
\setlength\belowcaptionskip{-0pt}
  \begin{tabular}{|p{1.7cm}|p{0.9cm}||p{0.8cm}|p{0.9cm}|p{1.cm}|p{1.cm}| p{1.cm}||p{0.8cm}|p{0.9cm}|p{1.cm}| p{1.cm}| p{1.cm}|}
  \hline
  \multirow{3}{*}{Lake}&\multirow{3}{*}{\makecell{Number\\of\\steps}} & \multicolumn{5}{c||}{Linear motion model} & \multicolumn{5}{c|}{Mixed motion model} \\ \cline{3-12}
  & & \multirow{2}{*}{Method} & \multirow{2}{*}{\makecell{Runtime\\(s)}} & \multicolumn{3}{c||}{$RMSE(m)$} & \multirow{2}{*}{Method} & \multirow{2}{*}{\makecell{Runtime\\(s)}} & \multicolumn{3}{c|}{$RMSE(m)$} \\ \cline{5-7} \cline{10-12}
  & & & &x &y &z & &  &x &y &z \\ \hline
  Bde Maka Ska  & 50 & EKF & 90.15 &4.09 & 12.43 & 0.42  & \textbf{EKF} &90.13 & \textbf{2.25} & \textbf{5.01} & \textbf{0.53} \\ \hline
  Bde Maka Ska  & 50 &UKF & 52.33 & 3.07 & 3.26 & 0.40 &UKF& 53.92 & 4.07 & 9.02 & 0.55\\ \hline
  Bde Maka Ska  & 50 & PF & 252.14 & 1.38& 3.57 & 0.67 &PF& 279.03 & 4.18 & 9.56 & 1.32\\ \hline
  Bde Maka Ska  & 50 & \textbf{MPF} &94.02 & \textbf{2.88} & \textbf{2.32} & \textbf{0.74} & MPF & 103.70 & 4.18 & 3.90& 1.15 \\ \hline
  \hline
  Nokomis  &50 & EKF & 57.77 & 3.92 & 5.60 & 0.43 & EKF& 59.84 & 4.96 & 7.12 & 0.49\\ \hline
  Nokomis  & 50 &UKF & 29.72 & 3.05 & 3.98& 0.42& UKF & 29.47 & 12.74 & 12.70 & 0.46 \\ \hline
  Nokomis  & 50 &PF & 251.50 & 13.69 & 23.12 & 0.89 & PF &291.11 & 11.90 & 18.60 & 0.98 \\ \hline
  Nokomis  & 50 & \textbf{MPF} & 79.42 & \textbf{2.84} & \textbf{3.10} & \textbf{0.76} & \textbf{MPF} & 79.87 & \textbf{4.12} & \textbf{4.80} & \textbf{0.76}\\ \hline
  \hline
  Hiawatha  & 30 & EKF& 12.97 & 3.56 & 3.57 & 0.48 & EKF & 12.88 & 3.33 & 2.63 & 0.51 \\ \hline
  Hiawatha  & 30 & UKF & 6.95 & 3.07 & 2.93 & 0.53 & UKF &7.32 & 3.03 & 3.00 & 0.57 \\ \hline
  Hiawatha  & 30 & \textbf{PF} & 137.52 & \textbf{1.66} & \textbf{2.57} & \textbf{1.13} & \textbf{PF} & 152.90 & \textbf{1.15} & \textbf{2.33} & \textbf{1.15} \\ \hline
  Hiawatha  &30 & MPF& 37.82 & 3.10 & 3.51 & 0.97 & MPF & 42.73 & 3.33 & 3.47 & 0.90 \\ \hline
  \hline
  Harriet  & 30 &EKF & 37.03 & 5.03 & 7.20 & 0.63 & EKF & 35.80 & 5.98 & 9.40 & 0.63 \\ \hline
  Harriet  & 30 & UKF & 19.43& 3.09 & 3.09 & 0.59 & UKF & 18.01& 4.08 & 6.03 & 0.54 \\ \hline
  Harriet  & 30 & PF & 149.71 & 1.77 & 3.72 & 1.13 & \textbf{PF} & 176.66 & \textbf{1.62} & \textbf{2.84} & \textbf{1.16} \\ \hline
  Harriet  &30 & \textbf{MPF} & 51.71 & \textbf{2.93} & \textbf{2.42} & \textbf{1.10} & MPF &54.39 & 3.13 & 2.75 & 1.09 \\ \hline
  \hline
  Turtle  & 50 & EKF& 46.20 & 3.50 & 4.26 & 0.50 & EKF & 46.47 & 3.66 & 3.20 & 0.44 \\ \hline
  Turtle  & 50 & UKF & 24.81 & 3.00 & 3.19 & 0.47 & UKF & 25.04 & 3.16 & 5.38 & 0.44 \\ \hline
  Turtle  & 50 & \textbf{PF} & 213.86 & \textbf{2.10} & \textbf{3.03} & \textbf{1.21} & PF & 249.93 & 2.68 & 3.78 & 1.24 \\ \hline
  Turtle  & 50 & MPF & 45.31 & 2.82 & 2.75 & 0.92 & \textbf{MPF} & 48.24 & \textbf{3.09} & \textbf{3.25} & \textbf{0.92} \\ \hline
    \hline
  Howard  & 50 & EKF& 88.94 & 2.98 & 3.29 & 0.38 & EKF & 89.33 & 3.08 & 3.90 & 0.43 \\ \hline
  Howard  & 50 & UKF & 46.00 & 3.18 & 3.37 & 0.47 & UKF & 46.00 & 6.53 & 7.12 & 0.48 \\ \hline
  Howard  & 50 & \textbf{PF} & 236.41 & \textbf{2.21} & \textbf{2.52} & \textbf{1.21} & \textbf{PF} & 282.78 & \textbf{3.26} & \textbf{3.34} & \textbf{1.23} \\ \hline
  Howard  & 50 & MPF & 58.75 & 2.68 & 2.71 & 0.89 & MPF & 61.76 & 3.50 & 4.38 & 0.91 \\ \hline
    \hline
  Waverly  & 50 & EKF & 60.00 & 3.79 & 3.84 & 0.49 & EKF & 59.95 & 5.62 & 3.11 & 0.59 \\ \hline
  Waverly & 50 & UKF & 31.59 & 3.15 & 2.64 & 0.48 & UKF & 32.57 & 6.05 & 4.24 & 0.47 \\ \hline
  Waverly  & 50 & \textbf{PF} & 227.69 & \textbf{1.61} & \textbf{1.20} & \textbf{1.30} & \textbf{PF} & 227.26 & \textbf{3.00} & \textbf{1.64} & \textbf{1.25} \\ \hline
  Waverly  & 50 & MPF & 49.60 & 2.46 & 2.45 & 0.96 & MPF & 53.39 & 5.67 & 3.30 & 0.99 \\ \hline
    \hline
  Pulaski  & 50 & EKF& 72.03 & 3.02 & 3.20 & 0.39 & EKF & 71.81 & 3.54 & 3.34 & 0.49 \\ \hline
  Pulaski  & 50 & UKF & 37.51 & 3.08 & 3.11 & 0.48 & UKF & 37.48 & 4.55 & 4.06 & 0.49 \\ \hline
  Pulaski  & 50 & \textbf{PF} & 224.57 & \textbf{2.03} & \textbf{1.14} & \textbf{1.29} & \textbf{PF} & 278.01 & \textbf{2.68} & \textbf{2.86} & \textbf{1.30} \\ \hline
  Pulaski  & 50 & MPF & 53.87 & 3.00 & 3.00 & 0.91 & MPF & 56.99 & 4.18 & 3.30 & 1.05 \\ \hline
\end{tabular}
\vspace{-5mm}
\label{tb:resulteval}
\end{table*}

\subsubsection{Linear motion case}
The EKF performs the worst for most lakes, and the results are not reliable since the performance varies between lakes. The UKF does not always perform best, but it gives reliable and relatively accurate results with lower computational complexity. The PF shows the most accurate result for some lakes, but it performs worst for Lake Nokomis. This result is likely caused by the fact that Lake Nokomis has a symmetrical structure and fewer variations in height. The MPF generally gives accurate and reliable results, and it is computationally cheaper than the PF. Overall, the MPF is the most reliable and accurate filter based on the results of these experiments. It is worth mentioning that the UKF is a good option if an AUV does not have much computational power and high accuracy is not required. The PF can be used for localization if the bathymetry data has enough variation in height and has an asymmetrical structure, provided that an AUV has enough computational power.
\subsubsection{Nonlinear motion case}
As in the linear motion cases, the PF and MPF generally perform better than the EKF and UKF. The UKF performs poorly for some cases due to the high nonlinearity of the motion. The PF displays the same issue that it has in the linear motion cases: the estimations diverge when there are not enough variations in bathymetry data. Similar to the linear cases, the MPF gives reliable and accurate results overall.
\subsubsection{Discussion}
The MPF consistently shows reliable and accurate results for both the linear and mixed motion cases while maintaining relatively low computational costs.
If an AUV needs a well-rounded algorithm for the localization problem, the MPF is the best filter among the four filters. 
However, if the bathymetry data of a lake has enough variation and the task requires high accuracy, then the PF is a better option. 
Determining what constitutes \enquote{enough variation} is a complicated question, one which should be explored in future work.
While PF and MPF have high accuracy, they require memory to store the state of their estimated particles, making them computationally expensive.
The MPF is significantly less computationally demanding than the PF, but still requires more memory and processing power than the UKF.
Therefore, for an AUV with low computational power, the UKF could be the best filter for localization if the AUV's motion is mostly going to be linear. 

\section{Conclusion}
Using bathymetry data and the measurements from a single-beam sonar altimeter and a depth sensor, 
%we present four low-cost \todo{low computational cost?} underwater localization algorithms for AUVs using EKF, UKF, PF, and MPF,
we present four localization algorithms based on the EKF, UKF, PF, and MPF respectively. We evaluate the performance of each filter in various aquatic environments and with multiple robot motions. The results demonstrate the feasibility of the Bayesian filter-based algorithms for localizing an AUV with bathymetry information using two low-cost sensors.
The MPF-based localization generally performs best, both in terms of accuracy and computational cost. However, the UKF can be a good alternative to the PF and MPF at the expense of accuracy if an AUV mostly actuates linearly and has limited computational power. Additionally, the PF seems to be the most accurate in water bodies with sufficient terrestrial variations if an AUV possesses the necessary computational power. Future work will focus on evaluating the performance of the proposed algorithms with field tests and analyzing the failures of particle filters on simple bathymetry in more detail.

% In order to propose low-cost localization algorithm for AUVs, we present the underwater localization algorithms for lakes with the simulated measurements from a single beam sonar and pressure sensor, and lake bathymetry data. The algorithm was developed with EKF, UKF, PF, and MPF, and tested. Their computational complexities were analyzed as well. EKF, UKF, PF, and MPF are adopted to develop the algorithms and their computational complexities are analyzed. The results show that the MPF-based localization gives more reliable and accurate results in general. UKF can be a good replacements to PF and MPF when the AUV actuates linear motions and does not have enough computational capacity at the expense of some accuracy. Additionally, PF can be used to get the most accurate results when a lake has enough territorial variations and the AUV has enough computational power.

% \section*{Acknowledgments}
% We are thankful to Chelsey Edge, Marc Ho, and Jiawei Mo for their assistance, the Minnesota DNR for the bathymetry data, and the MnDRIVE Initiative and Minnesota Robotics Institute for supporting this research.
% , and Cameron Fabbri for comments that greatly improved the manuscript.

\bibliographystyle{abbrv}
\bibliography{citation}
\end{document}